\documentclass{article}
\usepackage{spconf}

\usepackage{includepackages}
\def\th{{\raise.5ex\hbox{\small th }}}
\usepackage[utf8]{inputenc} 

\title{Graph based Manifold Regularized Deep Neural Networks for Automatic Speech Recognition}

 \name{Vikrant Singh Tomar$^{1,2}$, Richard Rose$^{1,3}$}
 \address{$^1$ McGill University, Department of Electrical and Computer Engineering, Montreal, Canada\\
 	$^2$ Fluent.ai, Montreal, QC, Canada\\
 	$^3$ Google Inc., NYC, USA}

\begin{document}

\maketitle

\begin{abstract} 
Deep neural networks (DNNs) have been successfully applied to a wide variety of acoustic modeling tasks in recent years. These include the applications of DNNs either in a discriminative feature extraction or in a hybrid acoustic modeling scenario. Despite the rapid progress in this area, a number of challenges remain in training DNNs. This paper presents an effective way of training DNNs using a manifold learning based regularization framework. In this framework, the parameters of the network are optimized  to preserve underlying manifold based relationships between speech feature vectors while minimizing a measure of loss between network outputs and targets. This is achieved by incorporating manifold based locality
 constraints in the objective criterion of DNNs. Empirical evidence is provided to demonstrate that training a network with manifold constraints 
 preserves structural compactness in the hidden layers of the network. Manifold regularization is applied to train bottleneck DNNs for feature extraction in hidden Markov model (HMM) based speech recognition. The experiments in this work are conducted on the Aurora-2 spoken digits and the Aurora-4 read news large vocabulary continuous speech recognition tasks. The performance is measured in terms of word error rate (WER) on these tasks. It is shown that the manifold regularized DNNs result in up to 37\%  reduction in WER relative to standard DNNs.
\end{abstract}


\section{Introduction}
\label{sec.intro}
Recently there has been a resurgence of research in the area of deep neural networks 
(DNNs) for acoustic modeling in automatic speech recognition (ASR)~ \cite{Jaitly2012, Hinton2012, Sainath2011,Tuske2012,Bengio2013,Deng2013}.
Much of this research has been concentrated on techniques for regularization of 
the algorithms used for DNN parameter estimation~\cite{Dahl2013, Zeiler2013, Deng2013b}.   
At the same time, there has also been a great deal of research on graph based 
techniques that facilitate the preservation of local neighborhood relationships among 
feature vectors for parameter estimation in a number of application areas~\cite{Tomar2014, Belkin2006, He2002, Jansen2006}.
Algorithms that preserve these local relationships are often referred to as having the 
effect of applying manifold based constraints.
This is because subspace manifolds are generally defined to be low-dimensional, 
perhaps nonlinear, surfaces where locally linear relationships exist among points
along the surface. 
Hence, preservation of these local neighborhood relationships is often thought, 
to some approximation, to have the effect of preserving the structure 
of the manifold.
To the extent that this assumption is correct, it suggests that the manifold shape 
can be preserved without knowledge of the overall manifold structure. 

This paper presents an approach for applying manifold based constraints to the 
problem of regularizing training algorithms for DNN based acoustic models in ASR.
This approach involves redefining the optimization criterion in DNN training to 
incorporate the local neighborhood relationships among acoustic feature vectors. To this end, this work uses discriminative manifold learning (DML) based constraints. The resultant training mechanism is referred to here as manifold regularized DNN (MRDNN) training
and is described in Section~\ref{sup.mrdnn}.
Results are presented demonstrating the impact of MRDNN training on both the  
ASR word error rates (WERs) obtained from MRDNN trained models 
and on the behavior of the associated nonlinear feature space mapping. 

Previous work in optimizing deep network training includes approaches for pre-training of network parameters such as restricted Boltzmann machine (RBM) based generative pre-training~\cite{Hinton2006, Yu2010, Dahl2012}, 
layer-by-layer discriminative pre-training~\cite{Seide2011}, 
and stacked auto-encoder based pre-training~\cite{Vincent2010, Larochelle2009}. 
However, more recent studies have shown that, if there are enough observation vectors available for training models, the network pre-training has little impact on ASR performance \cite{Deng2013, Deng2013b}. 
Other techniques, like dropout with the use of rectified linear units 
(ReLUs) as nonlinear elements in place of sigmoid units in the hidden layers of the network
are also thought to have the effect of regularization on DNN training~\cite{Dahl2013, Zeiler2013}.


The manifold regularization techniques described in this paper are applied to estimate
parameters for a tandem DNN with a low-dimensional bottleneck hidden layer~\cite{Grezl2007}.
The discriminative features obtained from the bottleneck layer are input to a Gaussian mixture model (GMM)
based HMM (GMM-HMM) speech recognizer   \cite{Tuske2012, Hermansky2000}. This work extends \cite{Tomar2014Interspeech} by presenting an improved MRDNN training algorithm that results in reduced computational complexity and better ASR performance.
ASR performance of the proposed technique is evaluated on two well known speech-in-noise corpora.
The first is the Aurora-2 speech corpus, a connected digit task~\cite{Aurora2}, and the 
second is the Aurora-4 speech corpus, a read news large vocabulary continuous speech recognition (LVCSR) task~\cite{Aurora4}. Both speech corpora and the ASR 
systems configured for these corpora are described in Section~\ref{sec.exp}. 

An important impact of the proposed technique is the well-behaved internal feature representations associated 
with MRDNN trained networks.
It is observed that the modified objective criterion results in a feature space in the internal network layers such that the local neighborhood relationships between feature vectors are preserved. 
That is, if two input vectors, $\B{x}_i$ and $\B{x}_j$, lie within the same local neighborhood in the input space, 
then their corresponding mappings, $\B{z}_i$ and $\B{z}_j$, in the internal layers will also be neighbors. 
This property can be characterized by means of an objective measure referred to as the contraction ratio \cite{Rifai2011},
which describes the relationship between the sizes of the local neighborhoods in the input and the mapped feature spaces. The performance of MRDNN training in producing mappings that preserve these local neighborhood relationships 
is presented in terms of the measured contraction ratio in Section~\ref{sup.mrdnn}. The locality preservation constraints associated with the MRDNN training are also shown in Section~\ref{sup.mrdnn} to lead to a more robust gradient estimation in error back propagation training. 

The rest of this paper is structured as follows. Second \ref{sec.backg}  
provides a review of basic principals associated with DNN training, manifold learning, and manifold regularization. 
Section \ref{sup.mrdnn} describes the MRDNN algorithm formulation, and provides a discussion of the contraction ratio as a measure of locality preservation in the DNN. Section \ref{sec.exp} describes the task domains, system configurations and presents the ASR WER performance results. Section \ref{sec.discuss} discusses  computational complexity of the proposed MRDNN training and the effect of noise on the manifold learning methods. In Section \ref{sec.mlpt}, another way of applying the manifold regularization to DNN training is discussed, where a manifold regularization is used only for first few training epochs. Conclusions and suggestions for future work are presented in Section \ref{sec.conclusion}.

\section{Background}
\label{sec.backg}
This section provides a brief review of DNNs and manifold learning principals as context for the MRDNN approach presented in Section \ref{sup.mrdnn}. Section \ref{sec.dnn} provides a summary of DNNs and their applications in ASR. An introduction to manifold learning and related techniques is provided in Section \ref{sec.ml}; this includes the discriminative manifold learning (DML) framework and locality sensitive hashing techniques for speeding up the neighborhood search for these methods. Section \ref{sec.mr} briefly describes the manifold regularization framework and some of its example implementations.

\subsection{Deep Neural Networks}
\label{sec.dnn}
A DNN is simply a feed-forward artificial neural network that has multiple hidden layers between the input and output layers. Typically, such a network produces a mapping $f_{dnn}: \B{x} \rightarrow \B{z}$ from the inputs, $\B{x}_i, i = 1 \dots N$, to the output activations, $\B{z}_i, i = 1 \dots N$. This is achieved by finding an optimal set of weights, $\B{W}$,  to minimize a global error or loss measure, $V(\B{x}_i, \B{t}_i, f_{dnn})$, defined between the outputs of the network and the targets, $\B{t}_i, i = 1 \dots N$. In this work,  $L2$-norm regularized DNNs are used as baseline DNN models. The objective criterion for such a model is defined as
\beq \footnotesize
\mathcal{F}( \B{W}) = \frac{1}{N} \sum_{i= 1}^N V(\B{x}_i, \B{t}_i, f_{dnn}) + \gamma_1 ||\B{W}||^2,
\label{eq.ebp}
\eeq
where the second term refers to the $L2$-norm regularization applied to the weights of the network, and $\gamma_1$ is the regularization coefficient that controls the relative contributions of the two terms.

The weights of the network are updated for several epochs over the training set using  mini-batch gradient descent based error back-propagation (EBP),
\beq \footnotesize
w^l_{m,n} \longleftarrow w^l_{m,n} + \eta \nabla_{w^l_{m,n}} \mathcal{F}( \B{W}),
\label{eq.weightupdate}
\eeq
where $w^l_{m,n}$ refers to the weight on the input to the  $n\th$ unit in the $l\th$ layer of the network from the $m\th$ unit in the preceding layer. The parameter $\eta$ corresponds to the gradient learning rate. The gradient of the global error with respect to $w^l_{m,n}$ is given as
\beq \footnotesize
\nabla_{w^l_{m,n}} \mathcal{F}( \B{W}) = - \Delta^l_{m,n} - 2 \gamma_1 w_{m,n}^l,
\label{eq.dnn_gradient}
\eeq
where $\Delta_{m,n}^l $ is the error signal in the $l\th$ layer and its form depends on both the error function and the activation function. Typically, a soft-max nonlinearity is used at the output layer along with the cross-entropy objective criterion. Assuming that the input to  $p\th$ unit in the final layer is calculated as $net_p^L = \sum_n w_{n,p}^L z_p^{L-1}$, the error signal for this case is given as $\Delta^L_{n,p}  = (z_p - t_p) z_{n}^{L-1}$ \cite{Dunne1997, Golik2013}.


\subsection{Manifold Learning}
\label{sec.ml}

Manifold learning techniques work under the assumption that high-dimensional data can be considered as a set of geometrically related points lying on or close to the surface of a smooth low dimensional manifold \cite{Tenenbaum1998,Saul2003,He2002}. Typically, these techniques are used for dimensionality reducing feature space transformations with the goal of preserving manifold domain relationships among data vectors in the original space \cite{Liu2011,Tenenbaum1998}.

The application of manifold learning methods to ASR is supported by the argument that speech is produced by the movements of loosely constrained articulators \cite{Jansen2006, Jansen2005}. Motivated by this, manifold learning can be used in acoustic modeling and feature space transformation techniques to preserve the local relationships between feature vectors.  This can be realized by embedding the feature vectors into one or more partially connected graphs \cite{Yan2007}. An optimality criterion is then formulated based on the preservation of manifold based geometrical relationships. One such example is locality preserving projections (LPP)  for feature space transformations in ASR \cite{YunRose2008}. The objective function in LPP is formulated so that feature vectors close to each other in the original space are also close to each other in the target space.


\subsubsection{Discriminative Manifold Learning}
\label{sec.dml}
Most manifold learning techniques are  unsupervised and non-discriminative, for example, LPP. As a result, these techniques do not generally exploit the underlying inter-class discriminative structure in speech features. For this reason, features belonging to different classes might not be well separated in the target space. A discriminative manifold learning (DML) framework for feature space transformations in ASR is proposed in \cite{Tomar2014,Tomar2012ISSPA,Tomar2012Interspeech}. The DML techniques incorporate discriminative training into manifold based nonlinear locality preservation.  For a dimensionality reducing mapping $f_{\mathcal{G}}: \B{x} \in \mathbb{R}^D \rightarrow \B{z} \in \mathbb{R}^d$, where $d << D$, these techniques attempt to preserve the within class manifold based relationships in the original space while maximizing a criterion related to the separability between classes in the output space.

In DML, the manifold based relationships between the feature vectors are characterized by two separate partially connected graphs, namely intrinsic and penalty graphs \cite{Yan2007,Tomar2014}. If the feature vectors, $\B{X} \in \mathbb{R}^{N \times D}$, are represented by the nodes of the graphs, the intrinsic graph,  $\mathcal{G}_{int} = \{\B{X}, \B{\Omega}_{int}\}$, characterizes the within class or within manifold  relationships between feature vectors. The penalty graph, $\mathcal{G}_{pen} = \{\B{X}, \B{\Omega}_{pen}\}$, characterizes the relationships between feature vectors belonging to different speech classes. $\B{\Omega}_{int}$ and $\B{\Omega}_{pen}$ are known as the affinity matrices for the intrinsic and penalty graphs respectively and represent the manifold based distances or weights on the edges connecting the nodes of the graphs. The elements of these affinity matrices are defined in terms of Gaussian kernels as

\beq\footnotesize
\omega^{int}_{ij} = \begin{cases}
exp \left( \frac{-|| \B{x}_i - \B{x}_j||^2} {\rho} \right) &;~C( \B{x}_i) = C(\B{x}_j), e( \B{x}_i, \B{x}_j)=1\\
0 &;~\mbox{Otherwise}
\end{cases} 
\label{eq.weights_int_lpda}
\eeq

and
\beq
\footnotesize
\omega^{pen}_{ij} =  \begin{cases}
exp \left( \frac{-|| \B{x}_i - \B{x}_j||^2} {\rho} \right) &;~C( \B{x}_i) \neq C(\B{x}_j), e( \B{x}_i, \B{x}_j)=1\\
0 &;~\mbox{Otherwise}
\end{cases},
\label{eq.weights_pen_lpda}
\eeq
where $C( \B{x}_i)$ refers to the class or label associated with vector $\B{x}_i$ and  $\rho$ is the Gaussian kernel heat parameter. The function $e( \B{x}_i, \B{x}_j)$ indicates whether $\B{x}_j$ is in the neighborhood of  $\B{x}_i$. Neighborhood relationships between two vectors can be determined by  $k$-nearest neighborhood (kNN) search.

In order to design an objective criterion that minimizes manifold domain distances between feature vectors belonging to the same speech class while maximizing the distances between feature vectors belonging to different speech classes, a graph scatter measure is defined. This measure represents the spread of the graph or average distance between feature vectors in the target space with respect to the original space. For a generic graph $\mathcal{G} = \{\B{X}, \B{\Omega} \}$, a measure of the graph's scatter for a mapping $f_{\mathcal{G}}: \B{x} \rightarrow \B{z} $ can be defined as
\beq \footnotesize
	F_{\mathcal{G}}(\B{Z}) = \sum_{i, j} ||\B{z}_i  - \B{z}_j  ||^2 \omega_{ij}.
	\label{eq.ge-scatter}
\eeq
The objective criterion in discriminative manifold learning is designed as the difference of the intrinsic and penalty scatter measures,
\beq\footnotesize
\begin{split}
F_{\mathcal{G}_{diff} }(\B{Z}) & = \sum_{i,j} || \B{z}_i - \B{z}_j ||^2 \omega_{ij}^{int} - \sum_{i,j} || \B{z}_i - \B{z}_j ||^2 \omega_{ij}^{pen}, \\
									& =  \sum_{i,j} || \B{z}_i - \B{z}_j ||^2 \omega_{ij}^{diff},
\end{split}
\label{eq.dml}
\eeq
where $ \omega_{ij}^{diff} =  \omega_{ij}^{int} - \omega_{ij}^{pen}$.

DML based dimensionality reducing feature space transformation techniques have been reported to provide significant gains over conventional techniques such as linear discriminant analysis (LDA) and unsupervised manifold learning based LPP for ASR \cite{Tomar2014,Tomar2012ISSPA,Tomar2012Interspeech}. This has motivated the use of DML based constraints for regularizing DNNs in this work.

\subsubsection{Locality Sensitive Hashing}
\label{sec.lsh}
One key challenge in applying manifold learning based methods to ASR is that due to large computational complexity requirements, these methods do not directly scale to big datasets. This complexity originates from the need to calculate a pair-wise similarity measure between feature vectors to construct neighborhood graphs. This work uses locality sensitive hashing (LSH) based methods for fast construction of the neighborhood graphs as described in \cite{Tomar2013ICASSP-LSH,Tomar2013Interspeech}. LSH creates hashed signatures of vectors in order to distribute them into a number of discrete buckets such that vectors close to each other are more likely to fall into the same bucket \cite{Indyk1998, Datar2004}. In this manner, one can efficiently perform similarity searches by exploring only the data-points falling into the same or adjacent buckets.

\subsection{Manifold Regularization}
\label{sec.mr}
Manifold regularization is a data dependent regularization framework that is capable of exploiting the existing manifold based structure of the data distribution. It is introduced in \cite{Belkin2006}, where the authors have also presented manifold extended versions of regularized least squares and support vector machines algorithms  for a number of text and image classification tasks.

There has been several recent efforts to incorporate some form of manifold based learning into a variety of machine learning tasks.
In~\cite{Subramanya}, authors have investigated manifold regularization in single hidden layer multilayer perceptrons for a phone classification task. Manifold learning based semi-supervised embedding have been applied to deep learning for a hand-written character recognition task in~\cite{Weston2008}.  Manifold regularized single-layer neural networks have been used for image classification in \cite{Ratle2009}.
In spite of these efforts of applying manifold regularization to various application domains, similar efforts are not known for training deep models in ASR.


\section{Manifold Regularized Deep Neural Networks}
\label{sup.mrdnn}

The multi-layered  nonlinear structure of DNNs makes them capable of learning  the highly nonlinear relationships between speech feature vectors. This section proposes an extension of EBP based DNN training by incorporating the DML based constraints discussed in Section \ref{sec.dml} as a regularization term. The algorithm formulation is provided in Section \ref{sec.mrdnn}. The proposed training procedure emphasizes local relationships among speech feature vectors along a low dimensional manifold while optimizing network parameters. To support this claim, empirical evidence is provided in Section \ref{sec.contract} that a network trained with manifold constraints has a higher capability of preserving the local relationships between the feature vectors than one trained without these constraints.

\subsection{Manifold Regularized Training}
\label{sec.mrdnn}
This work incorporates locality and geometrical relationships preserving manifold constraints as a regularization term in the objective criterion of a deep network. These constraints are derived from a graph characterizing the underlying manifold of speech feature vectors in the input space. The objective criterion  for a  MRDNN network producing a mapping $f_{mrdnn}: \B{x} \rightarrow \B{z}$ is given as follows, 
\beq \footnotesize
\begin{split}
\mathcal{F}(\B{W; Z}) = \sum_{i= 1}^N \left\{\vphantom{ \sum_{j=1}^{k}}\right. \frac{1}{N} & V(\B{x}_i, \B{t}_i, f_{mrdnn}) +  \gamma_1 ||\B{W}||^2 \\
& \left. + \gamma_2 \frac{1}{k^2}  \sum_{j=1}^{k} || \B{z}_i - \B{z}_j ||^2 \omega_{ij}^{int} \right\},
\label{eq.dnn.ml.mse}
\end{split}
\eeq
where $V(\B{x}_i, \B{t}_i, f_{mrdnn})$ is the loss between a target vector $\B{t}_i$ and output vector $\B{z}_i$ given an input vector $\B{x}_i$; V($\cdot$) is taken to be the cross-entropy loss in this work. \B{W} is the matrix representing the weights of the network. The second term in \eqref{eq.dnn.ml.mse} is the $L2$-norm regularization penalty on the network weights; this helps in maintaining smoothness for the assumed continuity of the source space. The effect of this regularizer is controlled by the multiplier $\gamma_1$. The third term in \eqref{eq.dnn.ml.mse} represents manifold learning based locality preservation constraints as defined in Section \ref{sec.dml}. Note that only the constraints modeled by the intrinsic graph,  $\mathcal{G}_{int} = \{\B{X}, \B{\Omega}_{int}\}$, are included, and the constraints modeled by the penalty graph, $\mathcal{G}_{pen} = \{\B{X}, \B{\Omega}_{pen}\}$, are ignored. This is further discussed in Section \ref{sec.arch}. The scalar $k$ denotes the number of nearest neighbors connected to each feature vector, and $\omega_{ij}^{int}$ refers to the weights on the edges of the intrinsic graph as defined in \eqref{eq.weights_int_lpda}. The relative importance of the mapping along the manifold is controlled by the regularization coefficient $\gamma_2$.

This framework assumes  that the data distribution is supported on a low dimensional manifold and corresponds to a data-dependent regularization that exploits the underlying manifold domain geometry of the input distribution. By imposing manifold based locality preserving constraints  on the network outputs in \eqref{eq.dnn.ml.mse}, this procedure encourages a mapping where relationships along the manifold are preserved and different speech classes are well separated.
The manifold regularization term penalizes the objective criterion for vectors that belong to the same neighborhood in the input space but have become separated in the output space after projection.

The objective criterion given in \eqref{eq.dnn.ml.mse} has a very similar form to that of a standard DNN given in \eqref{eq.ebp}. The weights of the network are optimized using EBP and gradient descent,
\beq \footnotesize
{\nabla_{\B{W}} \mathcal{F}(\B{ W ; Z})} = \sum_{i}^{N} \frac{\partial \mathcal{F}(\B{W ; Z})}{\partial \B{z}_i} \frac{\partial \B{z}_i}{\partial \B{W}},
\eeq
where $\nabla_{\B{W}} \mathcal{F}(\B{ W ; Z})$ is the gradient of the objective criterion with respect to the DNN weight matrix $\B{W}$. Using the same nomenclature as defined in \eqref{eq.dnn_gradient}, the gradient with respect to the weights in the last layer is calculated as
\beq \footnotesize
\begin{split}
	\nabla_{w^L_{n,p}} & \mathcal{F}( \B{W; Z}) = - \Delta^L_{n,p} - 2 \gamma_1 w_{n,p}^L\\
	& - \frac{2 \gamma_2}{k^2} \sum_{j=1}^k \omega_{ij} (z_{(i),p} - z_{(j),p}) \left( \frac{\partial z_{(i),p}}{\partial w^L_{n,p}} -  \frac{\partial z_{(j),p}}{\partial w^L_{n,p}} \right),
\end{split}
\label{eq.dnn_gradient_mrdnn}
\eeq
where $z_{(i),p}$ refers to the activation of the $p\th$ unit in the output layer when the input vector is $\B{x}_i$. The error signal, $\Delta^L_{n,p}$,  is the same as the one specified in \eqref{eq.dnn_gradient}. 

\subsection{Architecture and Implementation}
\label{sec.arch}
The computation of the gradient in \eqref{eq.dnn_gradient_mrdnn} depends not only on the input vector, $\B{x}_i$, but also its neighbors, $\B{x}_j, j = 1, \dots, k$, that belong to the same class.  Therefore, MRDNN training can be broken down into two components. The first is the standard DNN component that minimizes a cross-entropy based error with respect to given targets. The second is the manifold regularization based component that focuses on penalizing a criterion related to  preservation of neighborhood relationships.

\begin{figure}[h]
 \includegraphics[width = 0.5\textwidth]{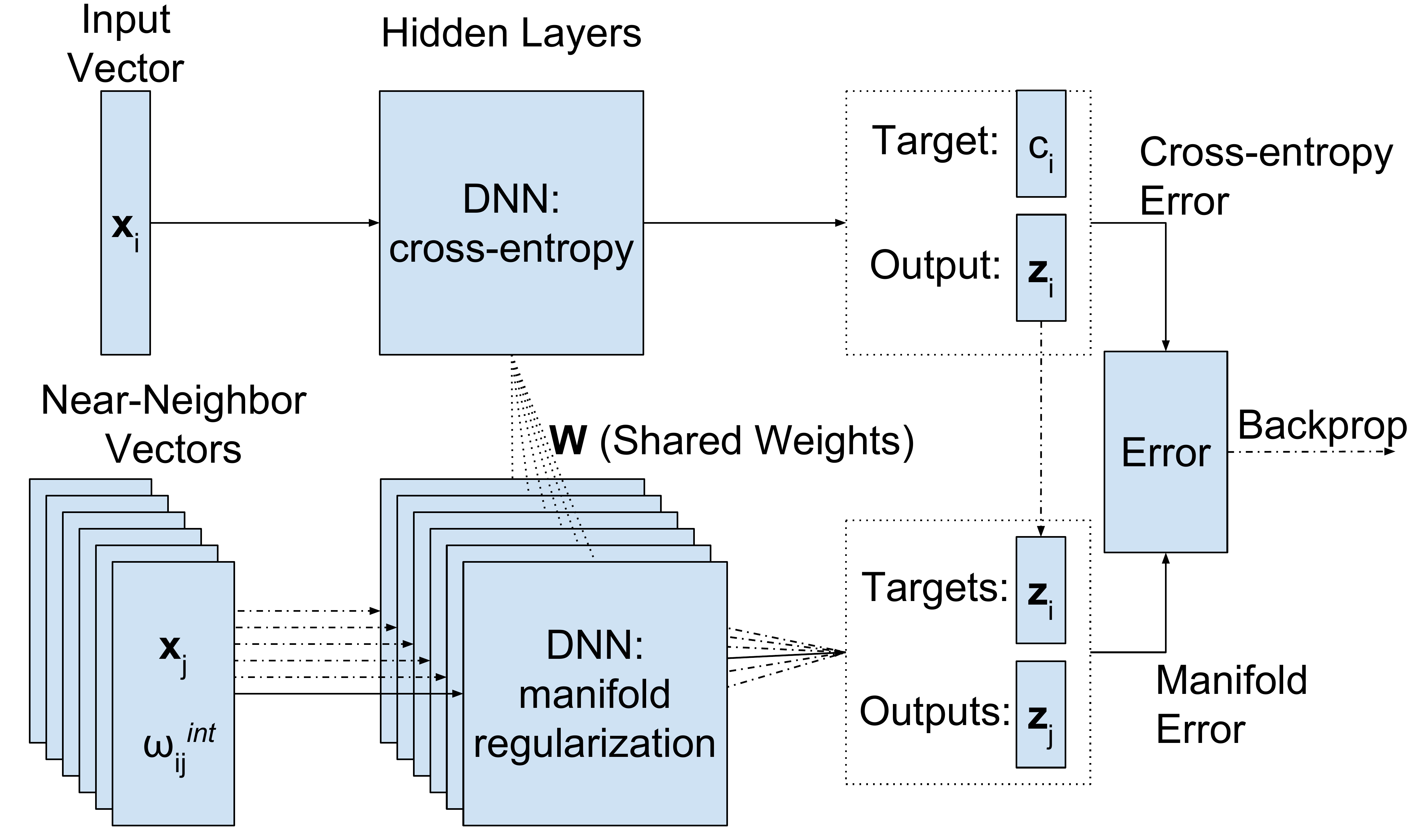}
 \caption{Illustration of MRDNN architecture. For each input vector, $\B{x}_i$, $k$ of its neighbors, $\B{x}_j$, belonging to the same class are selected and forward propagated through the network. For the neighbors, the target is set to be the output vector corresponding to $\B{x}_i$. The scalar $\omega_{ij}^{int}$ represents the intrinsic affinity weights as defined in \eqref{eq.weights_int_lpda}. }
 \label{fig.mrdnn_arch}
 \end{figure}

An architecture for manifold regularized DNN training is shown in Figure \ref{fig.mrdnn_arch}.  For each input feature vector, $\B{x}_i$, $k$ of its nearest neighbors, $\B{x}_j, j = 1, \dots, k$, belonging to the same class are selected.
These $k+1$ vectors are forward propagated through the network. 
This can be visualized as making $k+1$ separate copies of the DNN one for the input vector and the remaining $k$ for its neighbors. The DNN corresponding to the input vector, $\B{x}_i$, is trained to minimize cross-entropy error with respect to a given target, $t_i$. Each copy-DNN corresponding to one of the selected neighbors,  $\B{x}_j$, is trained to minimize a function of the distance between its output, $\B{z}_j$, and  that corresponding to the input vector, $\B{z}_i$. Note that the weights of all these copies are shared, and only an average gradient is used for weight-update.
This algorithm is then extended for use in mini-batch training.
 
It should be clear from this discussion that the relationships between a given vector and its neighbors play an important role
in  the weight update using EBP. Given the assumption of a smooth manifold,
the graph based manifold regularization is equivalent to penalizing rapid changes of the classification function.
This results in smoothing of the gradient learning curve that leads to robust computation of the weight updates.

In the applications of DML framework for feature space transformations, inclusion of both the intrinsic and  penalty graphs terms, as shown in \eqref{eq.dml}, is found to be important \cite{Tomar2012ISSPA, Tomar2014, Tomar2013ICASSP-NLPDA}. For this reason, experiments in the previous work included both these terms \cite{Tomar2014Interspeech}. However, in initial experiments performed on the corpora described in Section \ref{sec.exp}, the gains achieved by including the penalty graph were found to be inconsistent across datasets and different noise conditions. This may be because adding an additional term for discriminating between classes of speech feature vectors might not always impact the performance of DNNs since DNNs are inherently powerful discriminative models. Therefore, the penalty graph based term is not included in any of the experiments presented in this work, and the manifold learning based expression given in \eqref{eq.dnn.ml.mse} consists of the intrinsic component only.

\subsection{Preserving Neighborhood Relationships}
\label{sec.contract}
This section describes a study conducted to characterize the effect of applying manifold based constraints on the behavior of a deep network. This is an attempt to quantify how neighborhood relationships between feature vectors are preserved within the hidden layers of a manifold regularized DNN. To this end, a measure referred to as the contraction ratio is investigated. A form of this measure is presented in \cite{Rifai2011}. 

In this work, the contraction ratio is defined as the ratio of distance between two feature vectors at the output of the first hidden layer to that at the input of the network. 
The average contraction ratio between a feature vector and its neighbors can be seen as a measure of locality preservation and compactness of its neighborhood. Thus, the evolution of the average contraction ratio for a set of vectors as a function of distances between them can be used to characterize the overall locality preserving behavior of a network. To this end, a subset of feature vectors not seen during the training are selected. The distribution of pair-wise distances for the selected vectors is used to identify a number of bins. The edges of these bins are treated as a range of radii around the feature vectors. An average contraction ratio is computed as a function of radius in a range $r_1 < r \leq r_2$ over all the selected vectors and their neighbors falling in that range as
\beq
CR(r_1, r_2) = \frac{1}{N} \sum_{i=1}^N \sum_{j \in \Phi} \frac{1}{k_{\Phi}} \cdot \frac{||\B{z^1}_i - \B{z^1}_j||^2}{||\B{x}_i - \B{x}_j||^2},
\label{eq.contract}
\eeq
where for a given vector $\B{x}_i$, $\Phi$ represents a set of vectors $\B{x}_j$ such that $ r_1^2 < ||\B{x}_i - \B{x}_j||^2 \leq r_2^2$, and $k_{\Phi}$ is the number of such vectors. $\B{z^1}_i$ represents the output of the first layer corresponding to vector $\B{x}_i$ at the input. It follows that a smaller contraction ratio represents a more compact neighborhood. 

\begin{figure}[h]
	\includegraphics[width = 0.5\textwidth]{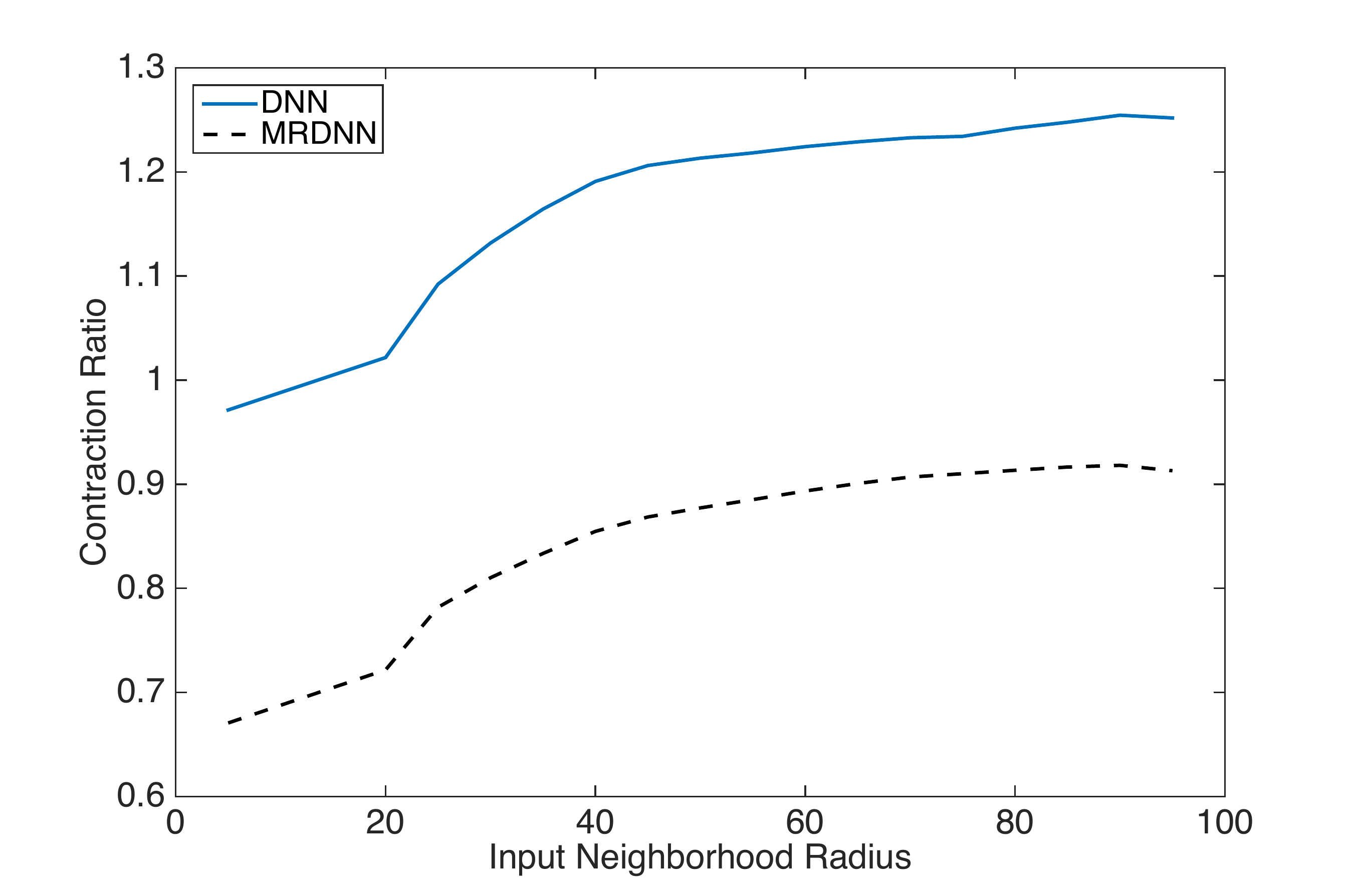}
	\caption{Contraction ratio of the output to input neighborhood sizes as a function of input neighborhood radius.}
	\label{fig.contractRatio}
\end{figure}

Figure \ref{fig.contractRatio} displays the contraction ratio of the output to input neighborhood sizes relative to the radius of the neighborhood in the input space for the DNN and MRDNN systems. The average contraction ratios between the input and the first layer's output features are plotted as functions of the median radius of the bins. It can be seen from the plots that the features obtained from a MRDNN  represent a more compact neighborhood than those obtained from a DNN. Therefore it can be concluded that the hidden layers of a MRDNN are able to learn the manifold based local geometrical representation of the feature space. It should also be noted that the contraction ratio increases with the radius indicating that the effect of manifold preservation diminishes as one moves farther from a given vector. This is in agreement with the local invariance assumption over low-dimensional manifolds.

\section{Experimental Study}
\label{sec.exp}
This section presents the experimental study conducted to evaluate the effectiveness of the proposed MRDNNs in terms of ASR WER.  The ASR performance of a MRDNN is presented and compared with that of DNNs without manifold regularization and the traditional GMM-HMM systems.  The experiments in this work are done on two separate speech-in-noise tasks, namely the Aurora-2, which is a connected digit task, and the Aurora-4, which is a read news LVCSR task. The speech tasks and the system setup are described in Section \ref{sec.task}. The results of the experiments on the Aurora-2 task are presented in Section \ref{sec.results_a2}. The results and comparative performance of the proposed technique on the Aurora-4 task are presented in Section \ref{sec.results_a4}. In addition, experiments on clean condition training sets of the Aurora-2 and Aurora-4 are also performed. The results for these experiments are provided in Section \ref{sec.cleantr_a2_a4}.

\subsection{Task Domain and System Configuration}
\label{sec.task}

The first dataset used in this work is the Aurora-2 connected digit speech in noise corpus. In most of the experiments, the Aurora-2 mixed-condition training set is used for training \cite{Aurora2}. Some experiments have also used the clean training set.  Both the clean and mixed-conditions training sets contain a total of 8440 utterances by 55 male and 55 female speakers. In the mixed-conditions set, the utterances are corrupted by adding four different noise types to the clean utterances. The conventional GMM-HMM ASR system is configured using the standard configuration specified in \cite{Aurora2}. This corresponds to using 10 word-based continuous density HMMs (CDHMMs) for digits 0 to 9 with 16 states per word-model, and additional models with 3 states for silence and 1 state for short-pause. In total, there are 180 CDHMM states each modeled by a mix of 3 Gaussians. During the test phase, five different subsets are used corresponding to uncorrupted clean utterances and utterances corrupted with four different noise types, namely subway, babble, car and exhibition hall, at signal-to-noise ratios (SNRs) ranging from 5 to 20 dB. There are 1001 utterances in each subset. The ASR performance obtained for the GMM-HMM system configuration agrees with those reported elsewhere \cite{Aurora2}.

The second dataset used in this work is the Aurora-4 read newspaper speech-in-noise corpus. This corpus is created by adding noise to the Wall Street Journal corpus \cite{Aurora4}. Aurora-4 represents a MVCSR task with a vocabulary size of 5000 words. This work uses the standard 16kHz mixed-conditions training set of the Aurora-4 \cite{Aurora4}. It consists of  {7138} noisy utterances from a total of 83 male and female speakers corresponding to about 14 hours of speech. One half of the utterances in the mixed-conditions training set are recorded with a primary Sennheiser microphone and the other half with a secondary microphone, which enables the effect of transmission channel. Both halves contain a mixture of uncorrupted clean utterances and noise corrupted utterances with the SNR levels varying from 10 to 20 dB in 6 different noise conditions (babble, street traffic, train station, car, restaurant and airport).  A bi-gram language model is used with a perplexity of 147. Context-depedent cross-word triphone CDHMM models are used for configuring the ASR system. There are a total of 3202 senones and each state is modeled by 16 Gaussian components.
Similar to the training set, the test set is recorded with the primary and secondary microphones. Each subset is further divided into seven subsets, where one subset is clean speech data and the remaining six are obtained by randomly adding the same six noise types as training at SNR levels ranging from 5 to 15 dB. Thus, there are a total of 14 subsets. Each subset contains 330 utterances from 8 speakers.

It should be noted that both of the corpora used in this work represent simulated speech-in-noise tasks. These are created by adding noises from multiple sources to speech utterances spoken in a quite environment. For this reason, one should be careful about generalizing these results to other speech-in-noise tasks.

For both corpora, the baseline GMM-HMM ASR systems are configured using 12-dimensional static Mel-frequency cepstrum coefficient (MFCC) features augmented by normalized $\log$ energy, difference cepstrum, and second difference cepstrum resulting in 39-dimensional vectors. The ASR performance is reported in terms of WERs. These GMM-HMM systems are also used for generating the context-dependent state alignments. These alignments are then used as the target labels for deep networks training as described below. 
  
Collectively, the term ``deep network" is used to refer to both DNN and MRDNN configurations. Both of these configurations include $L2$-norm regularization applied to the weights of the networks.  The ASR performance of these models is evaluated in a tandem setup where a bottleneck deep network is used as feature extractor for a GMM-HMM system. 
The deep networks take 429-dimensional input vectors that are created by concatenating 11 context frames of 12-dimensional MFCC features augmented with $\log$ energy, and first and second order difference. The number of units in the output layer is equal to the number of CDHMM states. The class labels or targets at the output are set to be the CDHMM states obtained by a single pass force-alignment using the baseline GMM-HMM system. The regularization coefficient $\gamma_1$ for the $L2$ weight decay is set to  0.0001. In the MRDNN setup, the number of nearest neighbors, $k$, is set to 10, and $\gamma_2$ is set to 0.001. While the manifold regularization techniques is not evaluated for hybrid DNN-HMM systems, one would expect the presented results to generalize to those models.
The experimental study performed in this work is limited to tandem DNN-HMM scenarios.  It is expected that the results reported here should also generalize to DNN training for hybrid DNN-HMM configurations. This is a topic for future work.

For the Aurora-2 experiments, the deep networks have five hidden layers. The first four hidden layers have 1024 hidden units each, and the fifth layer is bottleneck layer with 40 hidden units.  For the Aurora-4 experiments, larger networks are used. There are seven hidden layers. The first six hidden layers have 2048  hidden units each, and the seventh layer is bottleneck layer with 40 hidden units. This larger network for Aurora-4 is in-line with other recent work \cite{Su2015, Seltzer2013}. The hidden units use ReLUs as activation functions. The soft-max nonlinearity is used in the output layer with the cross-entropy loss between the outputs and targets of the network as the error \cite{Dunne1997, Golik2013}. After the EBP training, the 40-dimensional output features are taken from the bottleneck layer and de-correlated  using principal component analysis (PCA). Only features corresponding to the top 39 components are kept during the PCA transformation to match the dimensionality of the baseline system. The resultant features are used to train a GMM-HMM ASR system using a maximum-likelihood criterion. Although some might argue that PCA is not necessary after the compressed bottleneck output, in this work, a performance gain of 2-3\% absolute is observed with PCA on the same bottleneck features for different noise and channel conditions. No difference in performance is observed for the clean data.

\subsection{Results for the mixed-noise Aurora-2 Spoken Digits Corpus}
\label{sec.results_a2}

\begin{table}[t]
	\renewcommand{\arraystretch}{1.2}
	\footnotesize
	\caption{WERs for mixed noise training and noisy testing on the Aurora-2 speech corpus for GMM-HMM, DNN and MRDNN systems. The best performance has been highlighted for each noise type per SNR level.}
	\vspace{2ex}
	\centering
	\begin{tabular}{llllll} \hline
		\multirow{3}{*}{Noise}	& \multirow{3}{*}{Technique} &\multicolumn{4}{c}{SNR (dB)}  	\vspace*{1ex}\\
		\cline{3-6}
		& 	 		 	& 20  & 15 & 10 & 5\\
		\hline
		\multirow{3}{*}{Subway}	&  GMM-HMM  &	2.99 &	4.00	  &	6.21	& 11.89\\
		&  DNN 	 & 1.19 &	 {1.69}  & 2.95  &  6.02	\\
		&  MRDNN 		   &  \textbf{0.91} & \textbf{1.60 } & \textbf{2.39} & \textbf{5.67}  \\
		\hline
		\multirow{3}{*}{Babble} &  GMM-HMM    & 3.31 & 4.37  &	7.97	& 18.06\\
		&  DNN 		& 1.15 &	 {1.42}  & 2.81  &  7.38	\\
		& MRDNN		& \textbf{0.83} & \textbf{1.26} & \textbf{2.27} & \textbf{7.05}\\
		\hline
		\multirow{3}{*}{Car}	 &  GMM-HMM   &	2.77 &	3.36  &	5.45	& {12.31}\\
		&  DNN 	 & 1.05 &	 {1.85}  & 2.98  &  6.92	\\
		& MRDNN 	 & \textbf{0.84} & \textbf{1.37} & \textbf{2.59} & \textbf{6.38}\\
		\hline
		\multirow{3}{*}{Exhibition}	 &  GMM-HMM &	3.34 &	3.83	  &	6.64	& 12.72\\
		&  DNN 		& 1.23 &	 {1.54}  & 3.30  &  7.87	\\
		& MRDNN 			 & \textbf{0.96}  & \textbf{1.44} & \textbf{2.48} & \textbf{7.24} \\
		\hline
	\end{tabular}
	\label{tab.a2_relu}
\end{table}
The ASR WER for the Aurora-2 speech-in-noise task are presented in Table \ref{tab.a2_relu}. The models are trained on the mixed-noise set. The test results are presented in four separate tables each corresponding to a different noisy subset of the Aurora-2 test set. Each noisy subset is further divided into four subsets corresponding to noise corrupted utterances with 20 to 5dB SNRs. For each noise condition, the ASR results for features obtained from three different techniques are compared. The first row of each table, labeled `GMM-HMM', contains the ASR WER for the baseline GMM-HMM system trained using MFCC features appended with first and second order differences. The second row, labeled `DNN', displays results for the bottleneck features taken from the DNN described in Section \ref{sec.dnn}. The final row, labeled `MRDNN', presents the ASR WER results for the bottleneck  features obtained from the MRDNN described in Section \ref{sec.mrdnn}. For all the cases, the GMM-HMM and DNN configurations described in Section \ref{sec.task} are used. The initial learning rates for both systems are set to 0.001 and decreased exponentially with each epoch. Each system is trained for 40 epochs over the training set.

Two main observations can be made from the results presented in Table \ref{tab.a2_relu}. First, both DNN and MRDNN provide significant reductions in ASR WER over the  GMM-HMM system. The second observation can be made by comparing the ASR performance of DNN and MRDNN systems. It is evident from the results presented that the features derived from a manifold regularized network provide consistent gains over those derived from a network that is regularized only with the $L2$ weight-decay. The maximum relative WER gain obtained by using MRDNN over DNN is 37\%.

\subsection{Results for the mixed-noise Aurora-4 Read News Corpus}
\label{sec.results_a4}

The ASR WERs for the Aurora-4 task are given in Table \ref{tab.a4_relu}. All three acoustic models in the table are trained on the Aurora-4 mixed-conditions set. The table lists ASR WERs for four sets, namely clean, additive noise (noise), channel distortion (channel), and additive noise combined with channel distortion (noise + channel). These sets are obtained by grouping together the fourteen subsets described in Section \ref{sec.task}. The first row of the table, labeled `GMM-HMM', provides results for the traditional GMM-HMM system trained using MFCC features appended with first and second order differences. The second row, labeled `DNN', presents  WERs corresponding to the  features derived from the baseline $L2$ regularized  DNN system. ASR performances for the baseline systems reported in this work agree with those reported elsewhere \cite{Su2015,Seltzer2013}. The baseline set-up is consistent with that specified for the Aurora 4 task in order to be able to make comparisons with other results obtained in the literature for that task. The third row, labeled `MRDNN', displays the WERs for features obtained from a MRDNN. Similar to the case for Aurora-2, L2 regularization with the coefficient $\gamma_1$ set to  0.0001 is used for training both the DNN and MRDNN.
The initial learning rate is set to 0.003 and reduced exponentially for 40 epochs when the training is stopped.

Similar to the results for the Aurora-2 corpus, two main observations can be made by comparing the performance of the presented systems. First, both DNN and MRDNN provide large reductions in WERs over the  GMM-HMM for all conditions. Second, the trend of the WER reductions by using MRDNN over DNN is visible here as well.  The maximum relative WER reduction obtained by using MRDNN over DNN is 10.3\%.

\begin{table}[t]
	\footnotesize
	\centering
	\caption{ WER for mixed conditions training on the Aurora-4  task for  GMM-HMM, DNN, and MRDNN systems.}
	\begin{tabular}{lcccc} \hline
		{ } 	& 	 	{Clean}	&  {Noise} &  Channel & Noise + Channel\\
		\hline
		GMM-HMM   		& 13.02 &	18.62  & {20.27}	& 30.11  	\\
		
		\hline
		DNN   			&	5.91 &	10.32	  & {11.35}  & 22.78\\
		
		\hline
		MRDNN   			&	\textbf{5.30} &	\textbf{9.50}	  &  \textbf{10.11}  & \textbf{21.90} \\
		\hline
	\end{tabular}
	\label{tab.a4_relu}
	\vspace{-3ex}
\end{table}

The relative gain in the ASR performance for the Aurora-4 corpus is  less than that for the Aurora-2 corpus. This might be due to the fact that the performance of manifold learning based algorithms is highly susceptible to the presence of noise. This increased sensitivity is linked to the Gaussian kernel scale factor, $\rho$, defined in \eqref{eq.weights_int_lpda} \cite{Tomar2013ICASSP-NLPDA}. In this work, $\rho$, is set to be equal to $1000$ for all the experiments. This value is taken from previous work where it is empirically derived on a subset of Aurora-2 for a different task \cite{Tomar2012ISSPA, Tomar2013ICASSP-NLPDA}. While it is easy to analyze the effect of noise on the performance of the discussed models at each SNR level for the Aurora-2 corpus, it is difficult to do the same for the Aurora-4 corpus because of the way the corpus is organized. The Aurora-4 corpus only provides a mixture of utterances at different SNR levels for each noise type. 
Therefore, the average WER given for the Aurora-4 corpus might be affected by the dependence of ASR performance on the choice of $\rho$ and SNR levels, and a better tuned model is expected to provide improved performance. 
This is further discussed in Section \ref{sec.noise}.

Note that other techniques that have been reported to provide gains for DNNs were also investigated in this work. In particular, the use of RBM based generative pre-training was investigated to initialize the weights of the DNN system. Surprisingly, however, the generative pre-training mechanism did not lead to any reductions in ASR WER performance on these corpora. This is in agreement with other recent studies in the literature, which  have suggested that the use of ReLU hidden units on a relatively well-behaved corpus with enough training data obviates the need of a pre-training mechanism \cite{Deng2013b}.

\subsection{Results for Clean-condition Training}
\label{sec.cleantr_a2_a4}
In addition to the experiments on the mixed-conditions noisy datasets discussed in Sections \ref{sec.results_a2} and \ref{sec.results_a4}, experiments are also conducted to evaluate the performance of MRDNN for matched clean training and  testing scenarios. To this end, the clean training sets of the Aurora-2 and Aurora-4 corpora are used for training the deep networks as well as building manifold based neighborhood graphs \cite{Aurora2,Aurora4}. The results are presented in Table \ref{tab.cleantr_a2_a4}.  It can be observed from the results presented in the table that MRDNN provides 21.05\% relative improvements in WER over DNN for the Aurora-2 set and 14.43\% relative improvement for the Aurora-4 corpus.

\begin{table}[h]
\centering
	\renewcommand{\arraystretch}{1.2}
	\footnotesize
	\caption{WERs for clean training and clean testing on the Aurora-2 and Aurora-4 speech corpora for GMM-HMM, DNN, and MRDNN models. The last column lists \% WER improvements of MRDNN relative to DNN.}
	\vspace{2ex}
	\begin{tabular}{ccccc} \hline
		 & GMM-HMM  & DNN & MRDNN & \% Imp.	\\
		\hline

		  	Aurora-2	  & {0.93}  & {0.57} & {0.45} & 21.05 \\
		  			\hline
		  Aurora-4    	& 11.87   &  {8.11} & {6.94} & 14.43\\

		\hline
	\end{tabular}
	\label{tab.cleantr_a2_a4}
\end{table}

The ASR performance results presented for the Aurora-2 and the Aurora-4 corpora in Sections \ref{sec.results_a2}, \ref{sec.results_a4} and \ref{sec.cleantr_a2_a4} demonstrate the effectiveness of the proposed manifold regularized training for deep networks. While the conventional pre-training and regularization approaches did not lead to any performance gains, the MRDNN training provided consistent reductions in WERs. The primary reason for this performance gain is the ability of MRDNNs to learn and preserve the underlying low-dimensional manifold based relationships between feature vectors. This has been demonstrated with empirical evidence in Section \ref{sec.contract}. Therefore, the proposed MRDNN technique provides a compelling regularization scheme for training deep networks. 

\section{Discussion and Issues}
\label{sec.discuss}
There are a number of factors that can have an impact on the performance and application of MRDNN to ASR tasks. 
 This section highlights some of the factors and issues affecting these techniques.

\subsection{Computational Complexity}
\label{sec.complexity}
Though the inclusion of a manifold regularization factor in DNN training leads to encouraging gains in WER, it does so at the cost of additional computational complexity for parameter estimation. This additional cost for MRDNN training is two-fold. The first is the cost associated with calculating the  pair-wise distances for populating the intrinsic affinity matrix, $\Omega_{int}$. As discussed in Section \ref{sec.lsh} and \cite{Tomar2013ICASSP-LSH, Tomar2013Interspeech}, this computation complexity can be managed by using locality sensitive hashing for approximate nearest neighbor search without sacrificing ASR performance.

The second source of this additional complexity is the inclusion of $k$ neighbors for each feature vector during forward and back propagation. This results in an increase in the computational cost by a factor of $k$. This cost can be managed by 
 various parallelization techniques \cite{Sainath2011} and becomes negligible in the light of the massively parallel architectures of the modern processing units. This added computational cost is only relvent during the training of the networks and has no impact during the test phase or when the data is transformed using a trained network.  
 
 The networks in this work are training on Nvidia K20 graphics boards using tools developed on python based frameworks such as numpy, gnumpy and cudamat \cite{Mnih2009, Tieleman2010}. For DNN training, each epoch over the Aurora-2 set took 240 seconds and each epoch over the Aurora-4 dataset took 480 seconds. In comparison, MRDNN training took 1220 seconds for each epoch over the Aurora-2 set and 3000 seconds for each epoch over the Aurora-4 set.

\subsection{Effect of Noise}
\label{sec.noise}
In previous work, the authors have demonstrated that manifold learning techniques are very sensitive to the presence of noise \cite{Tomar2013ICASSP-NLPDA}. This sensitivity can be traced to the Gaussian kernel scale factor, $\rho$, used for defining the local affinity matrices in \eqref{eq.weights_int_lpda}. This argument might apply to MRDNN training as well. Therefore, the performance of a MRDNN might be affected by the presence of noise. This is visible to some extent in the results presented in Table \ref{tab.a2_relu}, where the WER gains by using MRDNN over DNN vary with SNR level. On the other hand, the Aurora-4 test corpus contains a mixture of noise corrupted utterances at different SNR levels for each noise type. Therefore, only an average over all SNR levels per noise type is presented.

There is a need to conduct extensive experiments to investigate this effect as was done in \cite{Tomar2013ICASSP-NLPDA}. However, 
if these models are affected by the presence of noise in a similar manner, additional gains could be derived by designing a method for building the intrinsic graphs separately for different noise conditions. During the DNN training, an automated algorithm could select a graph that best matches the estimated noise conditions associated with a given utterance. Training in this way could result in a MRDNN that is able to provide further gains in ASR performance in various noise conditions.

\subsection{Alternating Manifold Regularized Training}
\label{sec.mlpt}
Section \ref{sec.exp} has demonstrated reductions in ASR WER by forcing the output feature vectors to conform to the local neighborhood relationships present in the input data. This is achieved by applying the underlying input manifold based constraints to the DNN outputs throughout the training. There have also been studies in literature where manifold regularization is used only for first few iterations of model training. For instance, authors in \cite{Agarwal2010} have applied manifold regularization to multi-task learning. The authors have argued that optimizing deep networks by alternating between training with and without manifold based constraints can increase the generalization capacity of the model. 

Motivated by these efforts, this section investigates a scenario in which the manifold regularization based constraints are used only for the first few epochs of the training. 
All layers of the networks are  randomly initialized and then trained with the manifold based constraints  for first 10 epochs. The resultant network is then further trained for 20 epochs using the standard EBP without manifold based regularization. Note that contrary to the pre-training approaches in deep learning, this is not a greedy layer-by-layer training. 

ASR results for these experiments are given in Table \ref{tab.a2_relu_mlpt} for the Aurora-2 dataset. For brevity, the table only presents the ASR WERs as an average over the four noise types described in Table \ref{tab.a2_relu} at each SNR level. In addition to the average WERs for the DNN and MRDNN, a row labeled `MRDNN\_10' is given. This row refers to the case where manifold regularization is used only for first 10 training epochs. 

\begin{table}[t]
	\renewcommand{\arraystretch}{1.2}
	\footnotesize
	\caption{Average WERs for mixed noise training and noisy testing on the Aurora-2 speech corpus for DNN, MRDNN and MRDNN\_10 models.}
	\vspace{2ex}
	\begin{tabular}{lllllll} \hline
		\multirow{3}{*}{Noise}	& \multirow{3}{*}{Technique} &\multicolumn{5}{c}{SNR (dB)}  	\vspace*{1ex}\\
		\cline{3-7}
		& 	 		 	& clean & 20  & 15 & 10 & 5\\
		\hline
		\multirow{3}{*}{Avg.}	
		& DNN 		  & {0.91}  &  {1.16} & {1.63 } & {3.01} & {7.03}  \\
		&  MRDNN   & {0.66}  &  {0.88} & {1.42 } & {2.43} & {6.59}  \\
		&  MRDNN\_10		  & {0.62}  &  {0.81} & {1.37 } & {2.46} & {6.45}  \\
		\hline
	\end{tabular}
	\label{tab.a2_relu_mlpt}
\end{table}

%
%
%
%

A number of observations can be made from the results presented in Table \ref{tab.a2_relu_mlpt}. First, it can be seen from the results in Table \ref{tab.a2_relu_mlpt} that both MRDNN and MRDNN\_10 training scenarios improve ASR WERs when compared to the standard DNN training. Second, MRDNN\_10 training provides further reductions in ASR WERs for the Aurora-2 set. 
Experiments on the clean training and clean testing set of the Aurora-2 corpus also results in interesting comparison. In this case, the MRDNN\_10 training improved ASR performance to 0.39\% WER as compared to 0.45\% for MRDNN and 0.57\% for DNN training. That translates to 31.5\%  gain in ASR WER relative to DNNs. 

The WER reductions achieved in these experiements are  encouraging. Furthermore, this approach where manifold constraints are applied only for the first few epochs might lead to a more efficient manifold regularized training procedure because 
manifold regularized training has higher computational complexity than a DNN (as discussed in Section \ref{sec.complexity}). However, unlike \cite{Agarwal2010}, this work only applied one cycle of manifold constrained-unconstrained  training. It should be noted that these are preliminary experiments. Similar gains are not seen when these techniques are applied to the Aurora-4 dataset. Therefore, further investigation is required before making any substantial conclusions. 

\section{Conclusions and Future Work}
\label{sec.conclusion}
This paper has presented a framework for regularizing the training of DNNs using discriminative manifold learning based locality preserving constraints. The manifold based constraints are derived from a graph characterizing the underlying manifold of the speech feature vectors.  Empirical evidence has also been provided showing that the hidden layers of a MRDNN are able to learn the local neighborhood relationships between feature vectors. It has also been conjectured that the inclusion of a manifold regularization term in the objective criterion of DNNs results in a robust computation of the error gradient and weight updates. 

It has been shown through experimental results that the MRDNNs result in consistent reductions in ASR WERs that  range up to 37\%  relative to the standard DNNs on the Aurora-2 corpus. For the Aurora-4 corpus, the WER gains range up to 10\% relative on the mixed-noise training and 14.43\% on the clean training task. Therefore, the proposed manifold regularization based training can be seen as a compelling regularization scheme.

The studies presented here open a number of interesting possibilities. One would expect MRDNN training to have a similar impact in hybrid DNN-HMM systems. The application of these techniques in semi-supervised scenarios is also a topic of interest. To this end, the manifold regularization based techniques could be used for learning from a large unlabeled dataset, followed by fine-tuning on a smaller labeled set.

\section*{Acknowledgements}
The authors thank Nuance Foundation for providing financial support for this project. The experiments were conducted on the Calcul-Quebec and Compute-Canada supercomputing clusters. The authors are thankful to the consortium for providing the access and support.

\bibliographystyle{IEEEbib} 
\bibliography{IEEEabrv,library}      

\end{document}